\documentclass[10pt,twocolumn,letterpaper]{article}

\usepackage[pagenumbers]{cvpr} 
\usepackage[dvipsnames,table]{xcolor}

\usepackage{booktabs}
\usepackage{amsmath, bm, epstopdf, url, pifont, overpic, cases}
\usepackage{latexsym, amssymb, bbding, multirow, makecell,  diagbox, enumitem, caption}

\usepackage{arydshln}
\usepackage{lipsum}
\usepackage{anyfontsize}
\usepackage{t1enc}
\usepackage{pifont}

\usepackage{multirow}
\usepackage{adjustbox}
\usepackage{tabularx}
\usepackage{pifont}
\usepackage{color}

\newcommand{\R}[1]{\textcolor[rgb]{1.00,0.00,0.00}{#1}}
\newcommand{\B}[1]{\textcolor[rgb]{0.00,0.00,1.00}{#1}}

\usepackage{tabulary,overpic, makecell}
\usepackage{multicol}

\definecolor{Graylight}{gray}{0.9}

\usepackage{pifont}
\usepackage{adjustbox}
\usepackage[absolute,overlay]{textpos}
\definecolor{cvprblue}{rgb}{0,0,1}

\usepackage[pagebackref,breaklinks,colorlinks,citecolor=cvprblue]{hyperref}

\def\paperID{8245} 
\def\confName{CVPR}
\def\confYear{2024}

\title{\vspace{-0.5em}Uncertainty-Aware Source-Free Adaptive Image Super-Resolution with \\ Wavelet Augmentation Transformer\vspace{-0.5em}}

\author{Yuang Ai$^{1,2}$ \quad Xiaoqiang Zhou$^{1,3}$ \quad Huaibo Huang$^{1,2}$\thanks{Corresponding author.} \quad Lei Zhang$^{4,5}$ \quad Ran He$^{1,2,6}$ \\
$^{1}$MAIS \& CRIPAC, Institute of Automation, Chinese Academy of Sciences \\
$^2$School of Artificial Intelligence, University of Chinese Academy of Sciences\\
$^{3}$University of Science and Technology of China $^{4}$OPPO Research Institute\\
$^{5}$The Hong Kong Polytechnic University $^{6}$ShanghaiTech University\\
\tt\small shallowdream555@gmail.com, xq525@mail.ustc.edu.cn, \\ \tt\small huaibo.huang@cripac.ia.ac.cn, cslzhang@comp.polyu.edu.hk, rhe@nlpr.ia.ac.cn \\
\small{\url{https://shallowdream204.github.io/soda-sr/}}
\vspace{-1.5em}
}

\begin{document}
\maketitle
\begin{abstract}\vspace{-0.5em}
Unsupervised Domain Adaptation (UDA) can effectively address domain gap issues in real-world image Super-Resolution (SR) by accessing both the source and target data. Considering privacy policies or transmission restrictions of source data in practical scenarios, we propose a SOurce-free Domain Adaptation framework for image SR (SODA-SR) to address this issue, i.e., adapt a source-trained model to a target domain with only unlabeled target data. SODA-SR leverages the source-trained model to generate refined pseudo-labels for teacher-student learning. To better utilize pseudo-labels, we propose a novel wavelet-based augmentation method, named Wavelet Augmentation Transformer (WAT), which can be flexibly incorporated with existing networks, to implicitly produce useful augmented data. WAT learns low-frequency information of varying levels across diverse samples, which is aggregated efficiently via deformable attention. Furthermore, an uncertainty-aware self-training mechanism is proposed to improve the accuracy of pseudo-labels, with inaccurate predictions being rectified by uncertainty estimation. To acquire better SR results and avoid overfitting pseudo-labels, several regularization losses are proposed to constrain target LR and SR images in the frequency domain. Experiments show that without accessing source data, SODA-SR outperforms state-of-the-art UDA methods in both synthetic$\rightarrow$real and real$\rightarrow$real adaptation settings, and is not constrained by specific network architectures.
\end{abstract}    
\section{Introduction}
\label{sec:intro}

Single image super-resolution (SISR), which is a fundamental task in low-level vision, aims to reconstruct a high-resolution (HR) image from its low-resolution (LR) counterpart. In recent years, owing to the thriving advancements in deep learning, numerous deep learning-based approaches have been applied to SISR, culminating in significant breakthroughs in this task. Predominantly, these methods employ Convolutional Neural Networks (CNNs)~\cite{dong2015image, ledig2017photo} or Vision Transformers (ViTs)~\cite{Vit,liang2021swinir} as their architectural foundation. However, the majority are trained on synthetic datasets that generate LR images using simplistic and predetermined degradation kernels (\eg, bicubic). 

\begin{figure}[t]
\captionsetup{font=small}%
\scriptsize
\centering
\hspace{-38mm}
\includegraphics[width=0.45\textwidth]{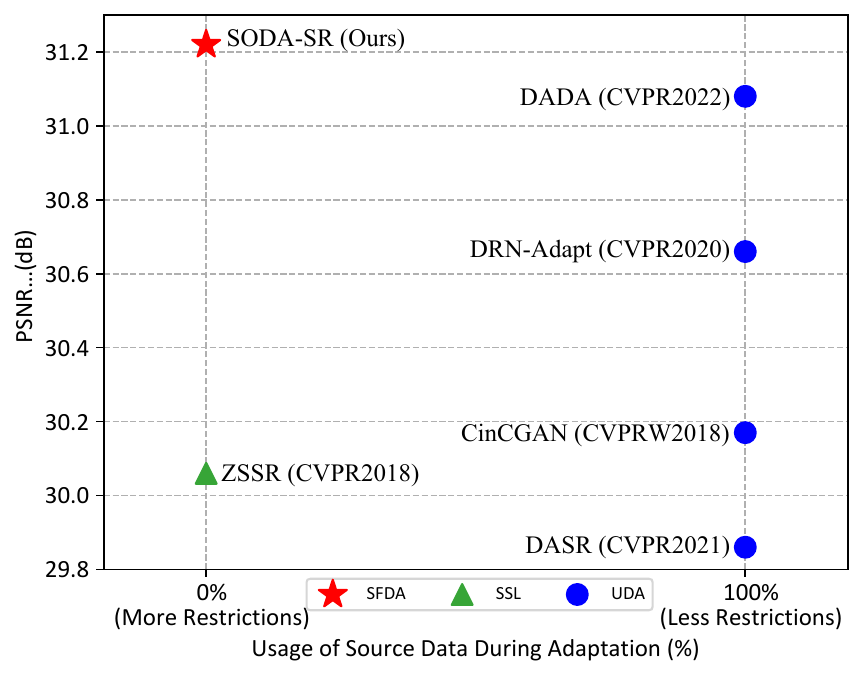}
\hspace{-68mm}\resizebox{.40\columnwidth}{!}{
\begin{tabular}[b]{lcc}
        \footnotesize
	\textbf{Method} & \textbf{Source Data} & \textbf{PSNR}\\
	\hline
    CinCGAN~\cite{CinCGAN} & \ding{52} & 30.17 \\
    DASR~\cite{DASR} & \ding{52} & 29.86 \\
	DRN-Adapt~\cite{DRN} & \ding{52} & 30.66  \\
	DADA~\cite{DADA} & \ding{52} & 31.08  \\
    \hline
    ZSSR~\cite{ZSSR} & \ding{56} & 30.06 \\ 
    \textbf{SODA-SR (Ours)} & \ding{56} & \textbf{31.22} \\
	\hline
	\multicolumn{3}{c}{\vspace{33.0mm}}
\end{tabular}
}
\vspace{-0.3cm}
\caption{PSNR vs. the usage of source data on the DRealSR~\cite{wei2020component} dataset. The less source data a method uses, the more restrictions it faces. SFDA and SSL represent source-free domain adaption and self-supervised learning methods respectively. } 
\label{fig:psnr_o2p}
\vspace{-0.7cm}
\end{figure}

However, the domain gap between the predetermined degradation and real-world degradation often leads to poor generalization capability of the SR model in real-world scenarios. To address this issue, several unsupervised domain adaptation (UDA) methods~\cite{CinCGAN,DASR,DRN,wang2021unsupervised,huang2022memory,huang2021memory} have been proposed to adapt the model from the source domain with synthetic image pairs to the target domain with unlabeled real images (\ie, synthetic$\rightarrow$real adaptation). Alternatively, some real-world datasets have been collected to train and evaluate real-world SISR methods, such as RealSR~\cite{cai2019toward} and DRealSR~\cite{wei2020component}. These datasets include LR and HR image pairs captured on the same scene taken by different cameras. In the meanwhile, there exists a significant gap between degradation kernels for images captured by different cameras, which can be regarded as a cross-device domain gap.~\cite{DADA} found that this kind of domain gap is harmful to the model's performance, and thus proposed a UDA method to adapt model from the source domain with paired real images to the target domain with unlabeled real images (\ie, real$\rightarrow$real adaptation). 

Though these UDA methods have achieved promising results in synthetic$\rightarrow$real and real$\rightarrow$real adaptation tasks, they do have certain limitations. Firstly, all these methods utilize source data to retain source knowledge and relieve domain shift during adaptation, which is often inaccessible due to privacy policies or transmission restrictions in practical scenarios. Besides, most of these methods are designed for specific SR network architectures and cannot be easily transferred to other networks, thus lacking generalizability.

In this paper, we propose and attempt to address a new and practical issue, namely Source-Free domain adaptation for image Super-Resolution (SFSR). It aims to adapt a model pre-trained on labeled source data to a target domain with only unlabeled target data.  Recently, several Source-Free Domain Adaptation (SFDA) methods have been proposed to address similar challenges in image classification~\cite{liang2020we,li2020model,liang2021source,litrico2023guiding}, semantic segmentation~\cite{liu2021source,fleuret2021uncertainty,lo2023spatio} and object detection~\cite{li2021free,li2022source,vs2023instance}. These methods are designed specifically for classification tasks with focus on obtaining reliable pseudo-labels or generating samples similar to the source domain distribution. However, when it comes to pixel-wise regression tasks such as image SR, which do not involve the notion of classes, these techniques are not directly applicable.

To address this issue, we present a novel method named SODA-SR, which is the first SOurce-free Domain Adaptation framework for image SR. Motivated by~\cite{semi_classification1,semi_classification2,semi_classification3} adding appropriate perturbations to the input (\eg, noise, data augmentation) or feature space (\eg, dropout~\cite{dropout}, stochastic depth~\cite{stochastic}) of the student model for better teacher-student learning in semi-supervised image classification, we adopt the teacher-student framework and apply the strategy of pseudo-labeling for optimization of the student model. However, existing perturbation methods designed for classification tasks cannot be directly used for SFSR, which may impact the performance of SR models. In light of this issue, we propose two distinct augmentation methods suitable for SR targeting the input and feature levels, respectively. Firstly, we flip and rotate the target LR image to generate seven geometrically augmented images for pseudo-label refinement.
Furthermore, we propose a novel Wavelet Augmentation Transformer (WAT) to implicitly generate augmented data, which can be flexibly incorporated with existing networks. By performing a multi-level wavelet decomposition for latent features, WAT learns low-frequency information of varying levels across diverse samples. It performs a proposed Batch Augmentation Attention (BAA) at different levels to mix image features batch-wisely and efficiently fuses these features through deformable attention~\cite{deformable}. WAT enables the student model with the ability to learn and explore appropriate augmentation in the feature space, which facilitates the student model in acquiring robust features.

Beyond that, an uncertainty-aware self-training mechanism is proposed to improve the accuracy of pseudo-labels by transferring knowledge from the target data to the teacher model. Specifically, the teacher model is updated with an exponential moving average of the student model to produce pseudo-labels. For one LR input, the teacher model runs multiple times to obtain the mean and variance as uncertainty estimation, which are then used to rectify pseudo-labels. Finally, we introduce several regularization losses to constrain the frequency information between target LR and SR images. These loss functions effectively prevent the student model from overfitting pseudo-labels by mining frequency information in target LR images, which leads to better SR results. As shown in Fig.~\ref{fig:psnr_o2p}, SODA-SR successfully generalizes the pre-trained source model on the target domain and achieves better PSNR against existing UDA and self-supervised methods.

The main contributions can be summarized as follows:
\begin{itemize}
    \item  
    We propose a novel SODA-SR framework to address the SFSR problem. To the best of our knowledge, this is the first research on SFSR.
    \item 
    We present a wavelet augmentation transformer (WAT) to implicitly synthesize augmented data. WAT learns cross-level low-frequency information of varying levels across diverse samples effectively and improves the robustness of the student model.
    \item 
    An uncertainty-aware self-training mechanism is introduced to improve the accuracy of pseudo-labels. Inaccurate predictions are rectified by uncertainty estimation. 
    \item 
    Extensive experiments show that our source-free SODA-SR outperforms state-of-the-art UDA methods and is not constrained by specific network architectures.
\end{itemize}

\section{Related Work}
\subsection{Single Image Super-Resolution}
With the rapid and dramatic development of deep learning, more and more SISR models have been proposed and yielded state-of-the-art performance among diverse datasets. Dong~\etal~\cite{dong2015image} presented an approach to learning the mapping function from LR images to HR images just using three convolutional layers. After that, a mass of CNN-based architectures~\cite{yulun_res_den,lim2017enhanced,wang2018esrgan,haris2018deep,zhang2018image,zhang2018residual,ledig2017photo,kim2016accurate,lai2018fast,huang2017wavelet,huang2019wavelet,duan2020cross} with more elaborate modules were proposed to improve the SISR performance. Recently, Transformer-based methods~\cite{chen2021pre,liang2021swinir,wang2022uformer,zamir2022restormer,zhang2022accurate,chen2023activating,li2023efficient,zhou2023msra,zhou2024ristra} were proposed for low-level vision tasks to utilize the great capability to model long-range dependency of Vision Transformer~\cite{Vit}. Some other works~\cite{saharia2022image,saharia2022palette,kawar2022denoising,wang2022zero,fei2023generative,ai2023multimodal} adopted diffusion models~\cite{ho2020denoising} to generate highly realistic images for image restoration.

\subsection{Real-World Image Super-Resolution}
Nowadays, in order to circumvent the limitations resulted from synthetic datasets, real-world SR has attracted more and more attention in the community. 
Some real-world SR datasets~\cite{chen2019camera,zhang2019zoom,cai2019toward,wei2020component} have been collected to train and evaluate real-world SR methods. Among them, DRealSR~\cite{wei2020component} stands as the singular real-world SR dataset encompassing multiple cameras and explicitly specifying the originating camera for each image.

In fact, there are relatively few UDA methods available for real-world SR. CinCGAN~\cite{CinCGAN} adopted a cycle-in-cycle network structure based on GAN to map real-world LR images to a noise-free space. DRN-Adapt~\cite{DRN} utilized paired synthetic data and unpaired real-world data to achieve adaptation with a dual regression constraint.  DASR~\cite{DASR} addressed the domain gap between training data and testing data with domain-gap aware training.~\cite{DADA} firstly explored the cross-device real-world SR and proposed an unsupervised mechanism to address this issue. However, all these methods need to access the source data and most of them are designed for specific SR network architectures. 
\subsection{Source-Free Domain Adaptation}
Recently, plenty of methods have been proposed to tackle SFDA for image classification, semantic segmentation and object detection. SHOT~\cite{liang2020we} and SHOT++~\cite{liang2021source} froze the classifier of the source model and matched the output target features to source feature distribution with information maximization and pseudo-label strategy. 
SFDA~\cite{liu2021source} generated fake samples with a BNS constraint and designed a dual attention distillation mechanism to transfer and retain the contextual information for semantic segmentation. LODS~\cite{li2022source} presented a style enhancement method to overlook the target domain style in source-free object detection. However, all of these methods are specifically designed for classification tasks and cannot be directly applied to pixel-level regression tasks, such as SR. To the best of our knowledge, this paper is the first work for  Source-Free domain adaptation for image Super-Resolution (SFSR).

\begin{figure}
    \captionsetup{font=small}%
    \centering
    \includegraphics[width=0.45\textwidth]{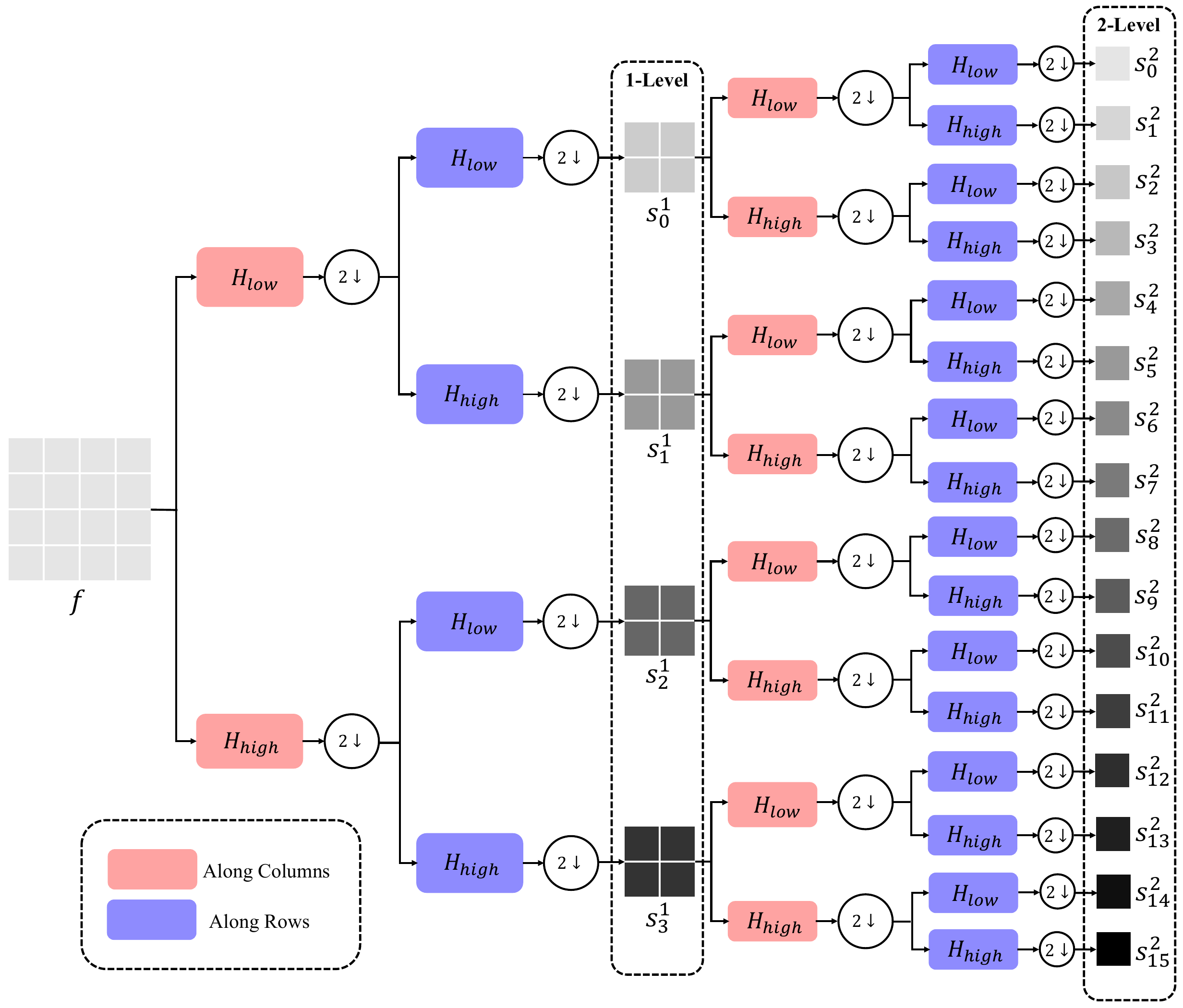}
    \vspace{-0.3cm}
    \caption{Illustration of 2-level haar wavelet packet transform (WPT). WPT employs low-pass filters $H_{low}$ and high-pass filters $H_{high}$ in a recursive manner to decompose the original features into multiple sub-bands at different frequency resolutions.}
    \vspace{-0.6cm}
    \label{fig:wavelet}
\end{figure}

\begin{figure*}[t]
  \centering
  \includegraphics[width=0.95\textwidth]{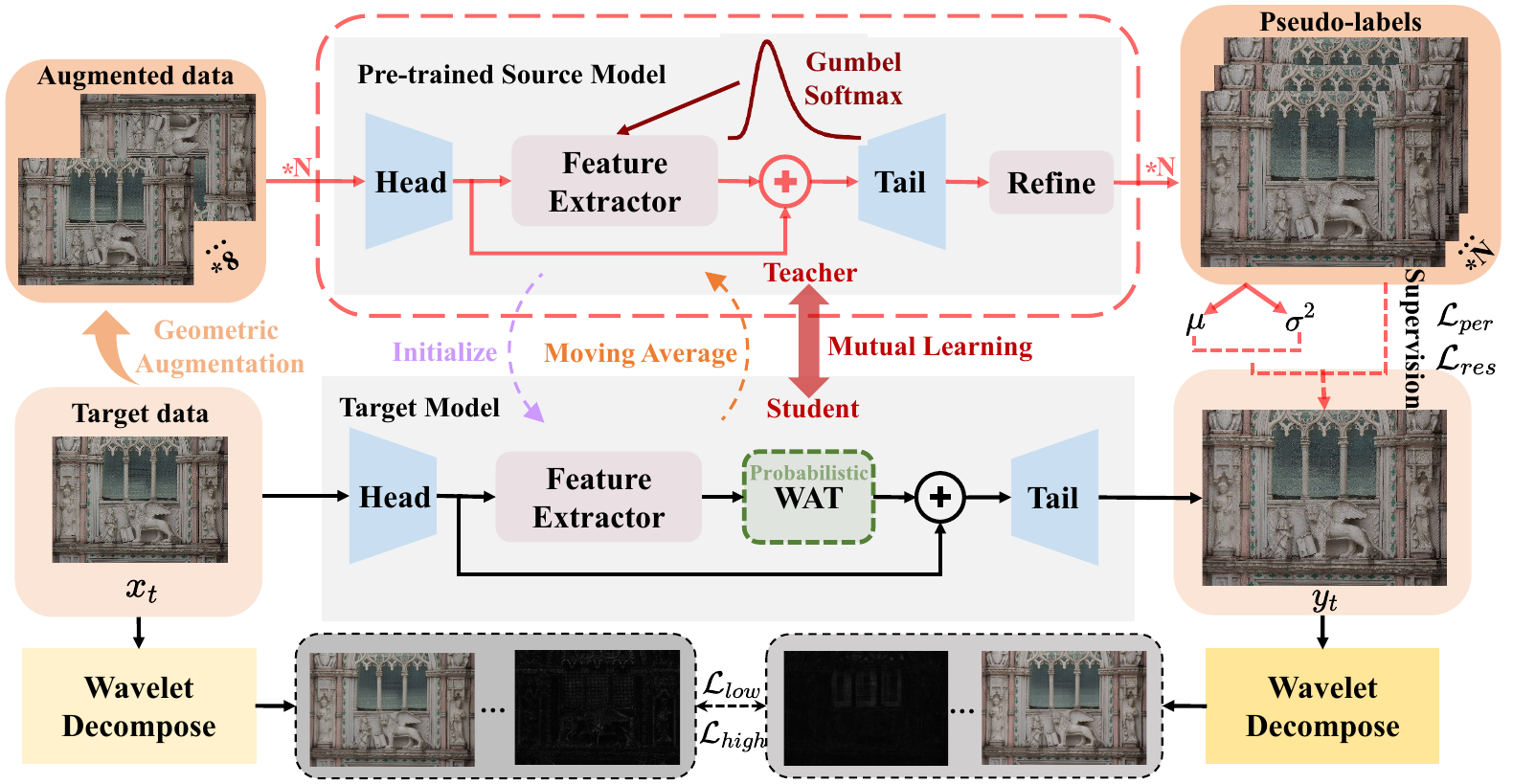}
  \vspace{-2pt}
 \caption{Architecture of the proposed  SODA-SR framework. One target LR input image together with its seven geometrically augmented images (\ie, rotate and flip the input) will be fed into the teacher model to generate the refined pseudo-label. The Softmax normalization function in the teacher model will be replaced by Gumbel-Softmax~\cite{gumbel}. For one LR input image, the teacher model will run multiple times to generate $N$ pseudo-labels and calculate their mean and variance for uncertainty estimation.
 }
  \vspace{-8pt}
  \label{fig:framework}
\end{figure*}
\section{Methodology}
According to the settings commonly used in the UDA task, the source dataset $D_s=\left \{\left (x_s^i,y_s^i  \right )   \right \}_{i=1}^{n_s} $ with $n_s$ pairs of labeled samples and the target dataset $D_t=\left \{x_t^i \right \}_{i=1}^{n_t}$ with $n_t$ unlabeled samples are given. In our source-free settings, the source dataset $D_s$ is only accessible during pre-training. Our goal is to adapt the pre-trained source model to the target domain without accessing the source data. 
\subsection{Overview}
As shown in Fig.~\ref{fig:framework}, SODA-SR is based on the teacher-student architecture. After pre-training, we can only access the well-trained teacher model $f_\xi$ and unlabeled target data $x_t$, where $\xi$ denotes the parameters of the teacher model. The student model $f_{\theta}$ has an additional wavelet augmentation transformer (WAT) built upon the teacher model. Let $\theta_w$ denote the parameters of WAT and $\theta_o$ denote the parameters of other modules in the student model, excluding WAT. $\theta_o$ is initialized as the pre-trained teacher model. The proposed WAT is based on wavelet-transform, more specifically wavelet packet transform (WPT). As illustrated in Fig.~\ref{fig:wavelet}, WPT can decompose the feature map into such sub-bands that have the same spatial size. Given a feature map of pixel-size $H \times W$ with a $\ell$-level WPT, we can get $4^{\ell}$ sub-bands of pixel-size $\frac{H}{2^{\ell}}\times \frac{W}{2^{\ell}}$.  WAT learns low-frequency information of varying levels using multi-level WPT.

SODA-SR consists of two distinct augmentation methods to facilitate teacher-student mutual learning. The first one is to rotate and flip one target LR image to generate seven geometrically augmented images. After feeding eight images into the teacher model to generate the SR images, the resulting SR images will be inverse transformed to their original geometry. The eight outputs will be averaged to produce the refined pseudo-label. The other is to use WAT to learn appropriate augmentation in the latent feature space. During training, there is a 50\% probability that the feature maps extracted by the feature extractor will be fed into the WAT or not. It's worth noting that WAT will not be utilized during inference, resulting in no additional computational cost.

\subsection{Wavelet Augmentation Transformer}
\begin{figure}[t]
\captionsetup{font=small}%
\scriptsize
\centering
\includegraphics[width=0.46\textwidth]{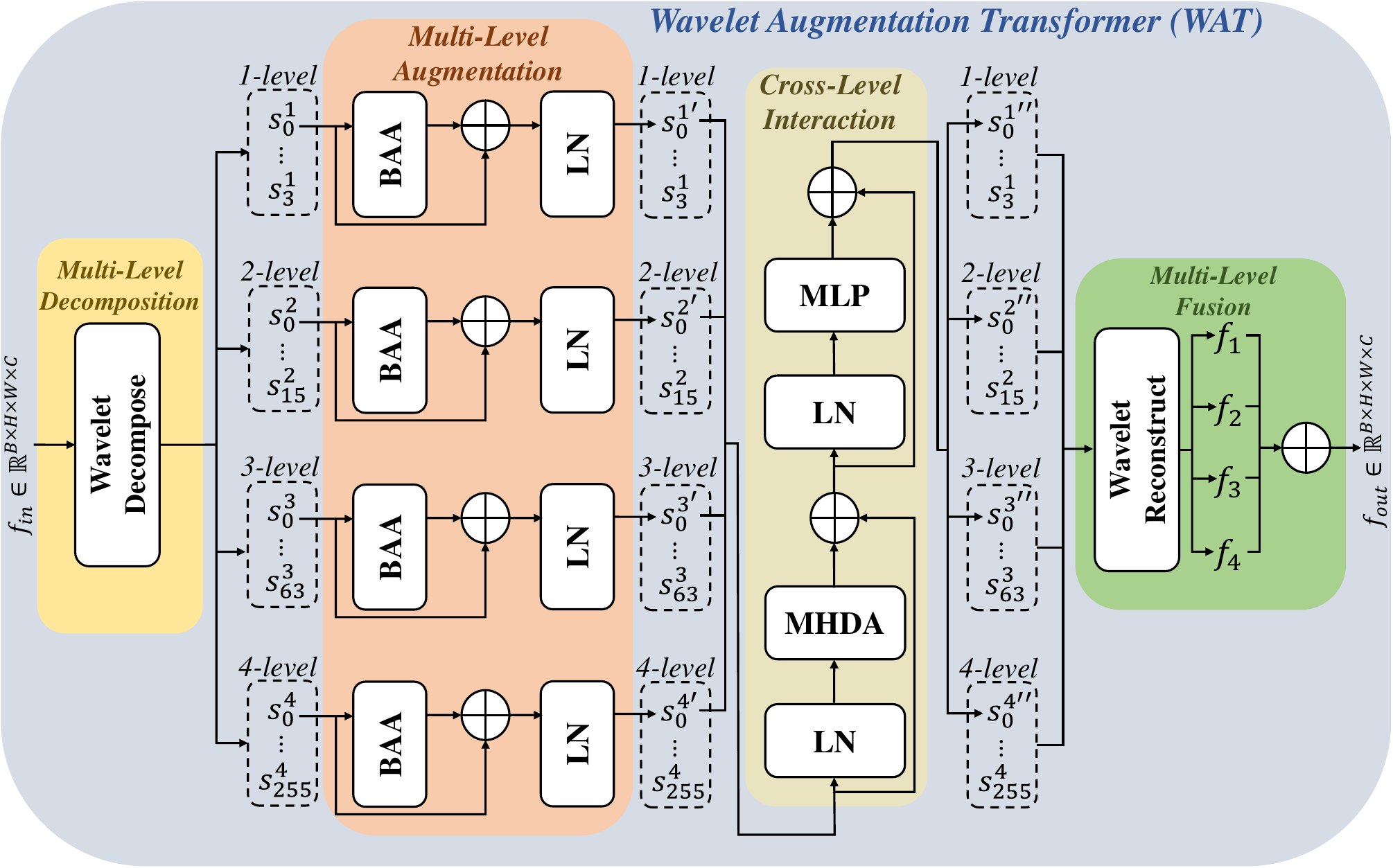}\label{fig:param_psnr}%
\caption{Wavelet Augmentation Transformer (WAT).} 
\label{fig:wat}
\vspace{-0.5cm}
\end{figure}

As shown in Fig.~\ref{fig:wat}, WAT consists of four key modules, including multi-level decomposition, multi-level augmentation, cross-level interaction, and multi-level fusion. We will proceed to introduce each of these modules as follows.

\textbf{Multi-Level Decomposition.} Given a batch of input image feature $f_{in}\in \mathbb{R}^{B\times H\times W\times C}$, WAT firstly employs a set of $\ell$-level WPT to decompose it into multi-level wavelet sub-bands, \ie, $\{s^{(\ell)}\in\mathbb{R}^{B\times m^{(\ell)}\times h^{(\ell)}\times w^{(\ell)}\times C}|\ell\in \mathrm{P}\}$, where $m^{(\ell)}=4^\ell, h^{(\ell)}=\frac{H}{2^\ell},w^{(\ell)}=\frac{W}{2^\ell}$ are the number, height, and width of the sub-bands, respectively. $\mathrm{P}=\{1,2,3,4\}$ is a set of wavelet levels. Then we flatten them on the spatial dimension and get $\{s^{(\ell)}\in\mathbb{R}^{B\times m^{(\ell)}\times n^{(\ell)}\times C}|\ell\in\mathrm{P}\}$, where $n^{(\ell)}=h^{(\ell)}\times w^{(\ell)}$. With the wavelet decomposition, the input image feature is transformed into the wavelet space, and the features are independent across different wavelet sub-bands. The image content is embedded in the low-frequency sub-band feature, while the detail and degradation information are embedded in the high-frequency sub-band feature. 

\textbf{Multi-Level Augmentation.} After the feature is disentangled in the wavelet feature space, we conduct the feature augmentation across different samples in the input batch. The proposed Batch Augmentation Attention (BAA) performs self-attention in a batch-wise manner. It's worth noting that the self-attention in BAA is conducted across the batch dimension rather than the spatial or channel dimension, i.e., computing cross-covariance across samples to achieve a learnable feature-level augmentation implicitly. To preserve the valuable high-frequency information in multi-level wavelet sub-bands, only the four low-frequency sub-bands, \ie, $\left \{ s^{(\ell)}_0 | \ell\in \mathrm{P} \right \}$ in Fig.~\ref{fig:wavelet}, will be fed into BAA simultaneously. Given $s^{(\ell)}_0\in \mathbb{R}^{B\times n^{(\ell)}\times C}$ as the input of the BAA. Firstly, we transpose the first two dimensions of $s^{(\ell)}_0$ so that $s^{(\ell)}_0 \in \mathbb{R}^{n^{(\ell)} \times B \times C}$. Then it employs the standard self-attention mechanism for $s^{(\ell)}_0$. We have $Q^{(\ell)},K^{(\ell)},V^{(\ell)} \in \mathbb{R}^{n^{(\ell)}\times B\times C}$and compute the output as
\begin{equation}
\mathrm{SA}(s^{(\ell)}_0) = \mathrm{softmax}(\frac{Q^{(\ell)}(K^{(\ell)})^T}{\sqrt{C}})V^{(\ell)}\in\mathbb{R}^{n^{(\ell)}\times B\times C},
 \end{equation}
where $(\cdot)^T$ represents the transpose of the second and third dimensions. Finally, we transpose the first two dimensions of the output. Then the final output of BAA has the same shape as the original $s^{(\ell)}_0\in\mathbb{R}^{B\times n^{(\ell)}\times C}$. 

BAA is akin to a feature-level Mixup~\cite{zhang2017mixup,hou2022batchformer}, enabling the interaction of information between distinct samples. The proposed multi-level augmentation module can learn low-frequency information of varying levels across diverse samples while also effectively preserving sensitive but valuable high-frequency information in image SR. 

\textbf{Cross-Level Interaction.} Inspired from~\cite{deformable}  that uses deformable attention to aggregate multi-scale feature maps in object detection, we employ the deformable attention to achieve cross-level information interaction. 

As shown in Fig.~\ref{fig:wat}, $\left \{ s^{(\ell)^\prime}_0 |\ell\in\mathrm{P} \right \}$ will be fed into the Multi-Head Deformable Attention (MHDA) module to facilitate information exchange across different levels, where $s^{(\ell)^\prime}_0\in\mathbb{R}^{B\times n^{(\ell)}\times C}$. Firstly we concatenate them on the second dimension, denoted as $X^{\prime}\in\mathbb{R}^{B\times N \times C}$, where $N=\sum_{\ell\in\mathrm{P}} n^{(\ell)}$. Let $x^{\prime}\in\mathbb{R}^{N\times C}$ denote one sample in $X^\prime$. For the $i^{\mathrm{th}}$ feature $x^\prime_i\in\mathbb{R}^{C}$ in $x^\prime$, where $i\in\{1,\cdots,N\}$, $K$ features are sampled in each level for each attention head. Let $p_i\in [0,1]^2$ be the normalized coordinates that represent the spatial position of $x^\prime_i$ in the original feature map. The position of sampling features can be denoted as $p_{hlik}= p_i+\Delta p_{h\ell ik}$, where $\Delta p_{h\ell ik}$ denotes the sampling offset of the $k^{\mathrm{th}}$ sampling point in the $h^{\mathrm{th}}$ attention head of the $\ell^{\mathrm{th}}$ level. Bilinear interpolation is used to sample the feature and the sampled feature is denoted as $x^\prime_{h\ell ik}$ for simplicity. The output of MHDA can be formulated as 
\begin{equation}
    \label{eq:deformable}
    x^{\prime \prime }_i = \sum_{h=1}^{H}W_h^{(1)}[\sum_{\ell\in P}\sum_{k=1}^{K}A_{h\ell ik}W_h^{(2)}x^\prime _{h\ell ik} ],
\end{equation}
where $H$ denotes the number of attention heads, $W_h^{(1)}\in\mathbb{R}^{C\times\frac{C}{H}}$ and $W_h^{(2)} \in \mathbb{R}^{\frac{C}{H}\times C}$ are projection matrices. $A_{h\ell ik}$ is the attention weight, which is obtained by projecting $x^\prime_i$ through a FC layer and normalizing it with softmax.

Through MHDA, the information across different levels can effectively interact. The LayerNorm layer is incorporated prior to both MHDA and MLP, with the addition of a residual connection for each module, which can be formulated as
\begin{equation}
\begin{split}
    X^{\prime\prime} &= \mathrm{MHDA}(\mathrm{LN}(X^\prime))+X^\prime,  \\
    X^{\prime\prime} &= \mathrm{MLP}(\mathrm{LN}(X^{\prime\prime}))+X^{\prime\prime},
\end{split}
\end{equation}
where $X^{\prime\prime}\in\mathbb{R}^{B\times N\times C}$. Then we split it on the second dimension and recover the multi-level feature maps $\left \{ s^{(\ell)^{\prime\prime}}_0 |\ell\in\mathrm{P}  \right \}$, whose information has been effectively aggregated via MHDA across different levels.

\textbf{Multi-Level Fusion.} As shown in Fig.~\ref{fig:wat}, we add up the four features after performing wavelet reconstruction on the wavelet sub-bands of different levels respectively. The output feature $f_{out}$ combines information of different levels. 

The proposed WAT performs BAA at different levels to mix image features batch-wisely and efficiently fuses these features through deformable attention. WAT can be regarded as a novel form of model noise~\cite{noisy_student}, which stimulates the student model to learn harder from pseudo-labels, thereby acquiring robust features.

\subsection{Uncertainty-aware Self-training Mechanism}
When the domain gap between the source domain and the target domain is huge, the pseudo-labels generated from the teacher model may be unreliable. To further improve the accuracy of the pseudo-labels, we present an uncertainty-aware self-training mechanism.

\textbf{Knowledge Transfer.} As shown in Fig.~\ref{fig:framework}, the parameters of the teacher model $\xi$ are updated with an exponential moving average (EMA) of the parameters of the student model (excluding WAT) $\theta_o$  after each training step:
\begin{equation}
\label{eq:EMA}
    \xi = \eta \cdot \xi + (1-\eta) \cdot \theta_o,
\end{equation}
where $\eta \in [0,1]$ is the decay rate, which is a hyper-parameter to control the update rate of the teacher model. This approach has been proven effective in semi-supervised learning~\cite{semi_classification1} and self-supervised learning~\cite{moco,grill2020bootstrap}. In our SODA-SR, the target domain knowledge learned by the student model can be slowly and progressively transferred to the teacher model via EMA, thereby improving the accuracy of pseudo-labels and promoting mutual learning between the teacher model and the student model.

\textbf{Pseudo-label Rectification.} 
In order to alleviate the adverse effects of inaccurate pseudo-labels, we incorporate uncertainty estimation into the self-training process to rectify the pseudo-labels.

Specifically, we replace the Softmax normalization function in the teacher model with Gumbel-Softmax~\cite{gumbel} to introduce stochasticity in the generation of pseudo-labels. Given 1D vector $v\in \mathbb{R}^n$, the output of Gumbel-Softmax is formulated as
\begin{equation}
    v_i=\frac{\mathrm{exp}((log(v_i)+g_i)/\tau)}{\sum_{j=1}^{n}\mathrm{exp}((log(v_j)+g_j)/\tau)}, 
    \label{eq:gumbel}
\end{equation}
where $g_1,\cdots,g_n$ are sampled from Gumbel$(0,1)$ distribution and $\tau$ is a temperature parameter. $g_1,\cdots,g_n$ introduce stochasticity to the teacher model, enabling it to produce diverse SR results.  For one target LR input, we run the teacher model multiple times to generate $N$ pseudo-labels $y_p^1, \cdots, y_p^N$. Then we compute the mean and variance as the uncertainty estimation, which is formulated as 
\begin{equation}
    y_{mean} = \frac{1}{N}\sum_{n=1}^{N}y_p^n, \ \sigma^2 = \frac{1}{N}\sum_{n=1}^{N}(y_p^n-y_{mean})^2.
    \label{eq:uncertain}
\end{equation}
Compared with existing methods, the proposed simple yet effective uncertainty estimation approach does not require additional components, \eg, Batch Normalization~\cite{teye2018bayesian} or Dropout~\cite{gal2016dropout}, which may affect the SR results. Then we compute the pixel-level confidence map $cof$ as following
\begin{equation}
    cof = \beta -\mathrm{Sigmoid}(\frac{\sigma ^2}{\alpha }),
    \label{eq:cof}
\end{equation}
where $\alpha$ and $\beta$ are hyper-parameters that adjust the value range of $cof$ and they are set empirically to 0.0004 and 1.5, respectively. The confidence map reflects the magnitude of pixel-wise uncertainty. During training, we calculate the pixel-wise weighted L1 loss between the output of the student model $f_\theta$ and the averaged pseudo-labels $y_{mean}$ using the confidence map:
\begin{equation}
    \mathcal{L}_{rec} = \|cof\odot f_\theta(x_t)-cof\odot y_{mean}  \|_1.
\end{equation}

The inaccurate pseudo-labels will be rectified by the confidence map. In addition to the L1 loss, we also utilize VGG-19~\cite{VGG} to calculate the perceptual loss $\mathcal{L}_{per}$.

\textbf{Regularization Losses.} Furthermore, to prevent the student model from overfitting pseudo-labels, two regularization losses are proposed to constrain the frequency information between LR and SR images. As shown in Fig.~\ref{fig:framework}, we conduct wavelet decomposition on LR and SR images at different levels, ensuring that the resulting wavelet sub-bands have the same resolution. We impose L1 loss in the low-frequency space while adversarial loss in the high-frequency space. The L1 loss is defined as 
\begin{equation}
\label{eq:loss_low}
    \mathcal{L}_{low}=\left \| \mathrm{wavelet}_\mathcal{L}^{(l1)}(x_t)-\mathrm{wavelet}_\mathcal{L}^{(l2)}(f_\theta(x_t)) \right \| _1,
\end{equation}
where $\mathrm{wavelet}_\mathcal{L}^{(l*)}(\cdot)$ represents the low-frequency sub-band of $l*$-level wavelet decomposition. The adversarial loss for the generator (\ie, the student model $f_\theta$) and the discriminator $\mathcal{D}$ is respectively defined as
\begin{equation}
    \mathcal{L}_{high}^G=-\mathbb{E}_{x_t}[log(\mathcal{D}(\mathrm{wavelet}_\mathcal{H}^{(l2)}(f_\theta(x_t))))],
\end{equation}
\begin{equation}
\begin{split}
    \mathcal{L}_{high}^D=&-\mathbb{E}_{x_t}[log(\mathcal{D}(\mathrm{wavelet}_\mathcal{H}^{(l1)}(x_t)))]\\
    &-\mathbb{E}_{x_t}[log(1-\mathcal{D}(\mathrm{wavelet}_\mathcal{H}^{(l2)}(f_\theta(x_t))))],
\end{split}
\end{equation}
where $\mathrm{wavelet}_\mathcal{H}^{(l*)}(\cdot)$ represents the high-frequency sub-band of $l*$-level wavelet decomposition.

\textbf{Full objective function.} The full objective function for the student model $f_\theta$ is defined as
\begin{equation}
\label{eq:loss_all}
\mathcal{L}=\mathcal{L}_{rec}+\lambda_1\mathcal{L}_{per}+\lambda_2\mathcal{L}_{low}+\lambda_3\mathcal{L}_{high}^G,
\end{equation}
where $\lambda_1$, $\lambda_2$ and $\lambda_3$ are loss weights to balance each item.

\section{Experiments}
\label{sec:ex}
\begin{table*}[t]
\vspace{0.3cm}
  \centering
  \renewcommand\arraystretch{0.82}
  \fontsize{8.8pt}{\baselineskip}\selectfont
  \setlength\tabcolsep{7pt}
  \begin{tabular}{ccccccccccc}
  \toprule\hline
  \multicolumn{1}{c|}{\multirow{+2}*{\textbf{Method}}} & \multicolumn{1}{c|}{\multirow{+2}*{\textbf{SF}}} & \multicolumn{3}{c|}{\textbf{{Panasonic} $\rightarrow$ {Sony}}} & 
  \multicolumn{3}{c|}{\textbf{{Sony} $\rightarrow$ {Panasonic}}} & 
  \multicolumn{3}{c}{\textbf{{Olympus} $\rightarrow$  {Panasonic}}} \\ \cline{3-11}
  \multicolumn{1}{c|}{\multirow{-2}*{\textbf{Method}}} & \multicolumn{1}{c|}{\multirow{-2}*{\textbf{SF}}} & 
  PSNR $\uparrow$ & SSIM $\uparrow$  &\multicolumn{1}{c|}{LPIPS $\downarrow$} & 
  PSNR $\uparrow$  & SSIM $\uparrow$   & \multicolumn{1}{c|}{LPIPS $\downarrow$}  & 
  PSNR $\uparrow$  & SSIM $\uparrow$    & LPIPS $\downarrow$              \\ \hline
  \rowcolor{gray!20}\multicolumn{11}{c}{\textit{Real$\rightarrow${}Real}}   \\ \hline
  \multicolumn{1}{c|}{\textit{Target Only}}
  & \multicolumn{1}{c|}{\ding{55}}
  & 32.71          
  & 0.855          
  & \multicolumn{1}{c|}{0.296}          
  & 32.33          
  & 0.845          
  & \multicolumn{1}{c|}{0.318}         
  & 32.33               
  & 0.845               
  & 0.318             
  \\
  \cdashline{1-11}[0.5pt/1.5pt]
  \multicolumn{1}{c|}{\textit{Source Only}} 
  & \multicolumn{1}{c|}{\ding{55}}  
  & 31.32          
  & 0.841          
  & \multicolumn{1}{c|}{{0.314}} 
  & 30.72          
  & 0.820         
  & \multicolumn{1}{c|}{0.372}          
  & 30.49               
  & 0.820               
  & 0.363              
  \\
  \multicolumn{1}{c|}{CinCGAN \cite{CinCGAN}}&\multicolumn{1}{c|}{\ding{55}} 
  & 27.76          
  & 0.821          
  & \multicolumn{1}{c|}{0.391}          
  & 28.33          & 0.792          
  & \multicolumn{1}{c|}{0.410}          
  & 29.37               
  & 0.799               
  & 0.381              
  \\
  \multicolumn{1}{c|}{DASR \cite{DASR}}&\multicolumn{1}{c|}{\ding{55}} 
  & 30.08          
  & 0.777          
  & \multicolumn{1}{c|}{\R{0.269}}          
  & 30.45          
  & 0.772         
  & \multicolumn{1}{c|}{\R{0.316}}          
  & 30.06               
  & 0.785               
  & \R{0.272}              
  \\
  \multicolumn{1}{c|}{DRN-Adapt \cite{DRN}}&\multicolumn{1}{c|}{\ding{55}}   
  & 31.85          
  & 0.845         
  & \multicolumn{1}{c|}{0.321}          
  & 30.96         
  & 0.821         
  & \multicolumn{1}{c|}{0.380}          
  & 30.80               
  & 0.822               
  & 0.356              
  \\
  \multicolumn{1}{c|}{DADA \cite{DADA}}  
  &\multicolumn{1}{c|}{\ding{55}}                           
  & \B{32.13} 
  & \B{0.849} 
  & \multicolumn{1}{c|}{0.327}          
  & \B{31.25} 
  & \B{0.825} 
  & \multicolumn{1}{c|}{{0.363}} 
  & \B{31.27}      
  & \B{0.824}      
  & {0.348}    
  \\
    \cdashline{1-11}[0.5pt/1.5pt]
    
  \multicolumn{1}{c|}{\textbf{SODA-SR (Ours)}} 
  &\multicolumn{1}{c|}{\ding{51}} 
  & \R{32.24} 
  &\R{0.851} 
  & \multicolumn{1}{c|}{\B{0.312}} 
  & \R{31.40} 
  &\R{0.833} 
  &\multicolumn{1}{c|}{\B{0.345}} 
  & \R{31.41} 
  &\R{0.832} 
  &\B{0.344} 
  \\ 
  \hline
  \rowcolor{gray!20}\multicolumn{11}{c}{\textit{Synthetic $\rightarrow$   Real}}                     \\ 
  \hline
  \multicolumn{1}{c|}{\textit{Source Only}}
  &\multicolumn{1}{c|}{\ding{55}}  
  & 31.44          
  & 0.828          
  & \multicolumn{1}{c|}{0.373}          
  & 30.45           
  & 0.808         
  & \multicolumn{1}{c|}{0.433}          
  & 30.44               
  & 0.806              
  & 0.434              
  \\
  \multicolumn{1}{c|}{CinCGAN  \cite{CinCGAN}}
  &\multicolumn{1}{c|}{\ding{55}}  
  & 27.59          
  & 0.788          
  & \multicolumn{1}{c|}{0.405}          
  & 27.19          
  & 0.743          
  & \multicolumn{1}{c|}{{0.414}} 
  & 28.38              
  & 0.739               
  & {0.422}     
  \\
  \multicolumn{1}{c|}{DASR  \cite{DASR}} 
 &\multicolumn{1}{c|}{\ding{55}}  
  & 29.95          
  & 0.764          
  & \multicolumn{1}{c|}{\R{0.298}}          
  & 29.79          
  & 0.749         
  & \multicolumn{1}{c|}{\R{0.339}} 
  & 30.02             
  & 0.777            
  & \R{0.293}    
  \\
  \multicolumn{1}{c|}{DRN-Adapt  \cite{DRN}} 
  &\multicolumn{1}{c|}{\ding{55}}  
  & 31.42        
  & 0.829         
  & \multicolumn{1}{c|}{0.359} 
  & 30.47 
  & 0.808        
  & \multicolumn{1}{c|}{0.429}   
  & 30.45             
  & \B{0.808}            
  & 0.433             
  \\
  \multicolumn{1}{c|}{DADA \cite{DADA}}    
 &\multicolumn{1}{c|}{\ding{55}}  
  & \B{31.50}
  & \B{0.830}
  & \multicolumn{1}{c|}{0.369}     
  & \B{30.72}       
  & \B{0.809}
  & \multicolumn{1}{c|}{0.376}      
  & \B{30.74}     
  & \B{0.808}     
  &  0.362           
  \\ 
    \cdashline{1-11}[0.5pt/1.5pt]

 \multicolumn{1}{c|}{\textbf{SODA-SR (Ours)}}    
 &\multicolumn{1}{c|}{\ding{51}}  
  & \R{31.61}
  & \R{0.831}
  & \multicolumn{1}{c|}{\B{0.354}}     
  & \R{30.80}       
  & \R{0.810}
  & \multicolumn{1}{c|}{\B{0.372}}      
  & \R{30.82}     
  & \R{0.809}     
  & \B{0.361}             
  \\
  \hline
  \multicolumn{1}{c|}{\multirow{2}{*}{\textbf{Method}}} & \multicolumn{1}{c|}{\multirow{+2}*{\textbf{SF}}}       & \multicolumn{3}{c|}{\textbf{{Panasonic}   $\rightarrow$ {Olympus}}}        & 
  \multicolumn{3}{c|}{\textbf{{Sony} $\rightarrow$ {Olympus}}}              & 
  \multicolumn{3}{c}{\textbf{{Olympus} $\rightarrow$ {Sony}}}        \\ \cline{3-11}
   \multicolumn{1}{c|}{\multirow{-2}{*}{\textbf{Method}}}    & \multicolumn{1}{c|}{\multirow{-2}*{\textbf{SF}}}                                     & PSNR $\uparrow$           & SSIM $\uparrow$           & \multicolumn{1}{c|}{LPIPS $\downarrow$}          & PSNR $\uparrow$           & SSIM $\uparrow$           & \multicolumn{1}{c|}{LPIPS $\downarrow$}          & PSNR $\uparrow$                & SSIM $\uparrow$                & LPIPS $\downarrow$              \\ \hline
  \rowcolor{gray!15}\multicolumn{11}{c}{\textit{Real$\rightarrow${}Real}}                                                                                                                                                                                                                         \\ \hline
  \multicolumn{1}{c|}{\textit{Target Only}}  
  &\multicolumn{1}{c|}{\ding{55}}  
  & 31.67          
  & 0.834          
  & \multicolumn{1}{c|}{0.375}          
  & 31.67         
  & 0.834          
  & \multicolumn{1}{c|}{0.375}          
  & 32.71              
  & 0.855           
  & 0.296             
  \\
  \cdashline{1-11}[0.5pt/1.5pt]
  \multicolumn{1}{c|}{\textit{Source Only}}        
  &\multicolumn{1}{c|}{\ding{55}}            
  & 30.23          
  & 0.812         
  & \multicolumn{1}{c|}{0.438}          
  & 30.48         
  & 0.810          
  & \multicolumn{1}{c|}{0.449}          
  & 30.45               
  & 0.812               
  & 0.323              
  \\
  \multicolumn{1}{c|}{CinCGAN  \cite{CinCGAN}}
  &\multicolumn{1}{c|}{\ding{55}} 
  & 28.85          
  & 0.791          
  & \multicolumn{1}{c|}{0.461}          
  & 30.17          
  & 0.814          
  & \multicolumn{1}{c|}{0.443}          
  & 30.05               
  & 0.823               
  & 0.365              
  \\
  \multicolumn{1}{c|}{DASR  \cite{DASR}}
  &\multicolumn{1}{c|}{\ding{55}} 
  & 29.32          
  & 0.768          
  & \multicolumn{1}{c|}{\R{0.306}}          
  & 29.86          
  & 0.762          
  & \multicolumn{1}{c|}{\R{0.372}}          
  & 30.29               
  & 0.787               
  & \R{0.270}              
  \\
  \multicolumn{1}{c|}{DRN-Adapt  \cite{DRN}} 
  &\multicolumn{1}{c|}{\ding{55}}  
  & 30.73          
  & 0.816         
  & \multicolumn{1}{c|}{0.431}          
  & 30.66        
  & 0.810          
  & \multicolumn{1}{c|}{0.459}          
  & 31.47               
  & 0.833              
  & 0.312    
  \\
  \multicolumn{1}{c|}{DADA \cite{DADA}}   
  &\multicolumn{1}{c|}{\ding{55}}                          
  & \B{31.08} 
  & \B{0.820} 
  & \multicolumn{1}{c|}{{0.433}} 
  & \B{31.08}
  & \B{0.817} 
  & \multicolumn{1}{c|}{{0.438}} 
  & \B{32.05}      
  & \B{0.843}      
  & 0.343 
  \\ 
  \cdashline{1-11}[0.5pt/1.5pt]
  
  \multicolumn{1}{c|}{\textbf{SODA-SR (Ours)}} 
  & \multicolumn{1}{c|}{\ding{51}}
  &\R{31.16} 
  &\R{0.828} 
  & \multicolumn{1}{c|}{\B{0.386}} 
  & \R{31.22} 
  & \R{0.828} 
  &\multicolumn{1}{c|}{\B{0.403}} 
  & \R{32.13} 
  &\R{0.850} 
  & \B{0.311}
  \\ 
  \hline
  \rowcolor{gray!15} \multicolumn{11}{c}{\textit{Synthetic $\rightarrow$   Real}}                       \\ 
  \hline
  \multicolumn{1}{c|}{\textit{Source Only}}      
  & \multicolumn{1}{c|}{\ding{55}}
  & 30.10        
  & 0.798        
  & \multicolumn{1}{c|}{0.480}    
  & 30.09       
  & 0.798   
  & \multicolumn{1}{c|}{0.473}     
  & 31.43           
  & 0.828            
  & 0.371              
  \\
  \multicolumn{1}{c|}{CinCGAN  \cite{CinCGAN}} 
  & \multicolumn{1}{c|}{\ding{55}}
  & 28.43      
  & 0.766        
  & \multicolumn{1}{c|}{0.407}  
  & 29.34        
  & 0.767      
  & \multicolumn{1}{c|}{0.451}     
  & 29.50        
  & 0.792           
  & 0.392            
  \\
  \multicolumn{1}{c|}{DASR  \cite{DASR}} 
    & \multicolumn{1}{c|}{\ding{55}}
  & 28.30       
  & 0.752       
  & \multicolumn{1}{c|}{\R{0.375}}      
  & 29.51      
  & 0.755        
  & \multicolumn{1}{c|}{\R{0.402}}  
  & 29.40           
  & 0.737            
  & \R{0.327}         
  \\
  \multicolumn{1}{c|}{DRN-Adapt  \cite{DRN}} 
    & \multicolumn{1}{c|}{\ding{55}}
  & 30.11        
  & 0.799   
  & \multicolumn{1}{c|}{0.475}      
  & 30.11        
  & 0.799     
  & \multicolumn{1}{c|}{0.473}   
  & 31.45           
  & \B{0.829}             
  & {0.362}    
  \\
  \multicolumn{1}{c|}{DADA \cite{DADA}}                  
  & \multicolumn{1}{c|}{\ding{55}}
  & \B{30.40} 
  & \B{0.800}
  & \multicolumn{1}{c|}{0.403}   
  & \B{30.62} 
  & \B{0.803} 
  & \multicolumn{1}{c|}{0.411}
  & \B{31.52}   
  & \B{0.829}    
  & 0.355            
  \\
    \cdashline{1-11}[0.5pt/1.5pt]
  
 \multicolumn{1}{c|}{\textbf{SODA-SR(Ours)}}                  
  & \multicolumn{1}{c|}{\ding{51}}
  & \R{30.42} 
  & \R{0.801}
  & \multicolumn{1}{c|}{\B{0.400}}   
  & \R{30.63} 
  & \R{0.804} 
  & \multicolumn{1}{c|}{\B{0.408}}
  & \R{31.54}   
  & \R{0.830}    
  & \B{0.351}            
  \\
  \hline
\bottomrule
  \end{tabular}
  \vspace{-4pt}
  \caption{Quantitative comparison with state-of-the-art UDA methods for $\times$ 4 SR. "\textbf{SF}" represents whether the method is under 
  source-free setting. Except \textit{Target Only}, the best and second best performance are in \R{red} and \B{blue} colors, respectively. }
  \vspace{-0.1cm}
  \label{tab:compare_uda}
\end{table*}

\begin{table*}[t]
\scriptsize
\center
\begin{center}
\resizebox{\linewidth}{!}{
\begin{tabular}{c|ccc|c|ccc|c|ccc}
    \toprule
         \multicolumn{4}{c|}{\textbf{{Panasonic Testset}}} &\multicolumn{4}{c|}{\textbf{{Sony Testset}}} &\multicolumn{4}{c}{\textbf{{Olympus Testset}}}
         \\
         \midrule
         Method & PSNR$\uparrow$ & SSIM$\uparrow$ & LPIPS$\downarrow$ & Method & PSNR$\uparrow$ & SSIM$\uparrow$ & LPIPS$\downarrow$ &  Method & PSNR$\uparrow$ & SSIM$\uparrow$ & LPIPS$\downarrow$
         \\
         \midrule
         MZSR~\cite{MZSR} & 28.73 & 0.785 & 0.398 & MZSR~\cite{MZSR}& 29.00 & 0.796 & 0.366 &MZSR~\cite{MZSR}& 28.54 & 0.777 & 0.443
         \\
         ZSSR~\cite{ZSSR} & 30.42 & 0.805 & 0.417 & ZSSR~\cite{ZSSR} & 31.34 & 0.823& 0.344 & ZSSR~\cite{ZSSR} & 30.06 & \B{0.794} & 0.455
         \\
         \textbf{Ours (S$\rightarrow$P)} & \B{31.40} & \R{0.833} & \B{0.345} & \textbf{Ours (O $\rightarrow$S)} &\B{32.13} & \B{0.850} & \R{0.311} & \textbf{Ours (P$\rightarrow$O)} & \B{31.16} & \R{0.828} & \R{0.386}
         \\
         \textbf{Ours (O$\rightarrow$P)} & \R{31.41} & \B{0.832} & \R{0.344} & \textbf{Ours (P$\rightarrow$S)} & \R{32.24} & \R{0.851} & \B{0.312} & \textbf{Ours (S$\rightarrow$O)} & \R{31.22} & \R{0.828} & \B{0.403}
         \\
         \bottomrule
    \end{tabular}}
\end{center}
\vspace{-5pt}
    \caption{Quantitative comparison with typical self-supervised SR methods for $\times4$ SR. "P", "S" and "O" represent Panasonic, Sony and Olympus, respectively. For the test set of one camera, our method has two adaptation settings.}
    \vspace{-0.35cm}
    \label{tab:compare_ssl}
\end{table*}

\subsection{Experimental Setup}
\noindent \textbf{Datasets and Metrics.} We evaluate our method on the DRealSR~\cite{wei2020component} dataset. DRealSR is a large-scale real-world SR benchmark, which is collected by five DSLR cameras (\ie, Panasonic, Sony, Olympus, Nikon, and Canon) in real-world scenarios. Following DADA~\cite{DADA}, we choose images captured from three cameras (Panasonic, Sony and Olympus) for our experiments, which contain 197, 145 and 190 image pairs for training; 20, 17 and 19 image pairs for testing, respectively. The image SR performance is evaluated by calculating PSNR, SSIM, and LPIPS~\cite{lpips}. PSNR and SSIM are computed on the Y channel (transformed YCbCr space) and RGB space, respectively.

\vspace{0.1cm}
\noindent \textbf{Implementation Details.}
Following DADA~\cite{DADA}, we adopt CDC~\cite{wei2020component} as the backbone network to achieve a fair comparison. 
Following DADA~\cite{DADA}, the training patch size is set to 48 $\times$ 48. We use the Adam optimizer with $\beta_1=0.9$, $\beta_2=0.999$ and a fixed learning rate of $2\times 10^{-6}$.  The decay rate $\eta$ in Eq.~(\ref{eq:EMA}) is set to 0.999 to make the training process more stable. The temperature parameter $\tau$ in Eq.~(\ref{eq:gumbel}) is set to 0.1 to mitigate the impact on the performance of the teacher model while introducing stochasticity. The number of generated pseudo-labels $N$ is set to 5. The levels of wavelet decomposition $l_1$ and $l_2$ in Eq.~(\ref{eq:loss_low}) are set to 1 and 3, respectively. For hyper-parameters in Eq.~(\ref{eq:loss_all}), the loss weights $\lambda_1,\lambda_2,\lambda_3$ are set to 0.01, 0.1, and 0.005, respectively. The proposed framework will converge after about 4000 iterations with a batch-size of 32.

\subsection{Comparison with state-of-the-art methods}
\begin{figure*}[!htbp]
	\captionsetup{font=small}
	\scriptsize
	\centering
	
	\newcommand{\h}{0.105}
	\newcommand{\wa}{0.12}
	\newcommand{\wb}{0.16}
	\newcommand{\g}{-0.7mm}
	\setlength\tabcolsep{1.8pt}
	\renewcommand{\arraystretch}{1}
	\resizebox{1.00\linewidth}{!} {
		\begin{tabular}{cc}
			
			\newcommand{\name}{fig/visual/}
			\renewcommand{\h}{0.110}
			\newcommand{\w}{0.176}
			\begin{tabular}{cc}
				\begin{adjustbox}{valign=t}
					\begin{tabular}{c}
						\includegraphics[height=0.262\textwidth, width=0.374\textwidth]{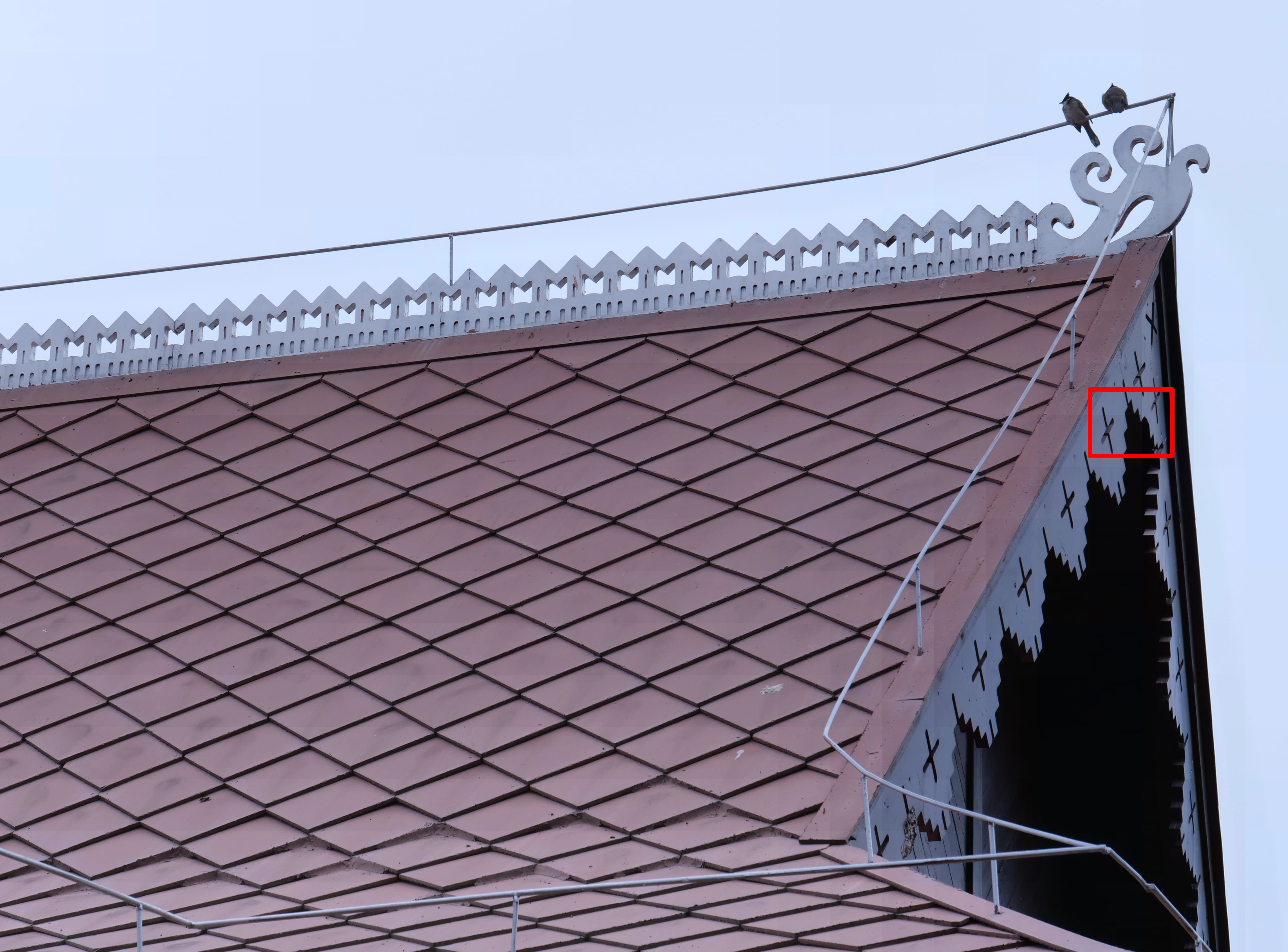}
						\\
						DRealSR ($4\times$): panasonic\_16
					\end{tabular}
				\end{adjustbox}
				\begin{adjustbox}{valign=t}
					\begin{tabular}{cccccc}
						\includegraphics[height=\h \textwidth, width=\w \textwidth]{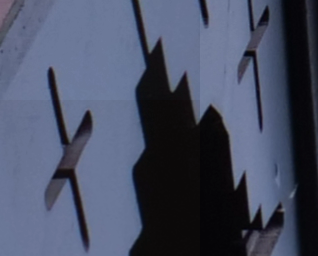} \hspace{\g} &
						\includegraphics[height=\h \textwidth, width=\w \textwidth]{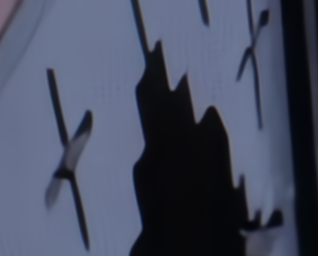} \hspace{\g} &
						\includegraphics[height=\h \textwidth, width=\w \textwidth]{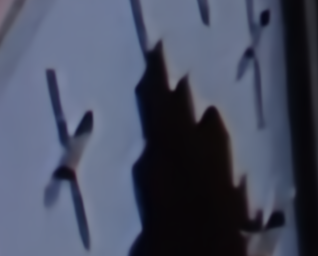} \hspace{\g} &
						\includegraphics[height=\h \textwidth, width=\w \textwidth]{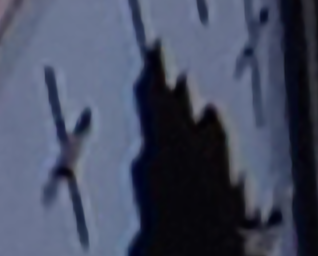} \hspace{\g} &
						\includegraphics[height=\h \textwidth, width=\w \textwidth]{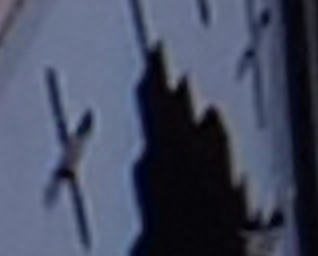} 
						\\
						HR \hspace{\g} &
						Target Only \hspace{\g} &
						Source Only \hspace{\g} &
						MZSR~\cite{MZSR} & 
                            ZSSR~\cite{ZSSR} \hspace{\g}
						\\
                            PSNR/SSIM \hspace{\g} &
                            32.74/0.9285 \hspace{\g} &
                            30.83/0.9040 \hspace{\g} &
                            27.79/0.8344 \hspace{\g} &
                            29.23/0.8486 \hspace{\g} 
                            \\
						\vspace{-2mm}
						\\						
						\includegraphics[height=\h \textwidth, width=\w \textwidth]{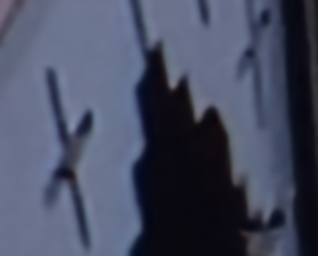} \hspace{\g} &
						\includegraphics[height=\h \textwidth, width=\w \textwidth]{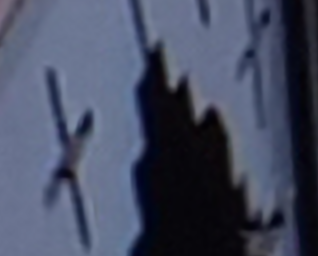} \hspace{\g} &
						\includegraphics[height=\h \textwidth, width=\w \textwidth]{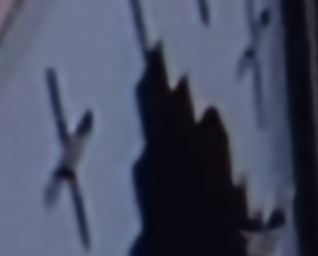}
						\hspace{\g} &		
						\includegraphics[height=\h \textwidth, width=\w \textwidth]{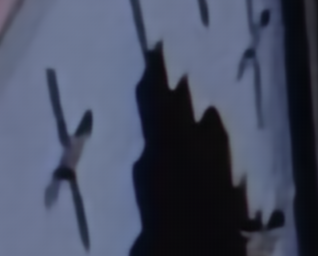} \hspace{\g} &
						\includegraphics[height=\h \textwidth, width=\w \textwidth]{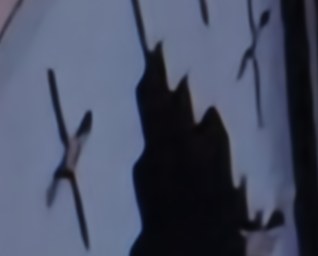} 
						\\ 
						
						CinCGAN~\cite{CinCGAN} \hspace{\g} &
						DASR~\cite{DASR}  \hspace{\g} &
						DRN-Adapt~\cite{DRN} \hspace{\g} &
						DADA~\cite{DADA}
						& \textbf{SODA-SR} (Ours)
						\\
                            29.17/0.8672 \hspace{\g} &
                            29.33/0.8522 \hspace{\g} &
                            29.32/0.8664 \hspace{\g} &
                            31.31/0.9082 \hspace{\g} &
                            31.74/0.9138 \\
					\end{tabular}
				\end{adjustbox}
			\end{tabular}
			
		\end{tabular}

 }

 \resizebox{1.00\linewidth}{!} {
		\begin{tabular}{cc}
			
			\newcommand{\name}{fig/visual/}
			\renewcommand{\h}{0.110}
			\newcommand{\w}{0.176}
			\begin{tabular}{cc}
				\begin{adjustbox}{valign=t}
					\begin{tabular}{c}
						\includegraphics[height=0.262\textwidth, width=0.374\textwidth]{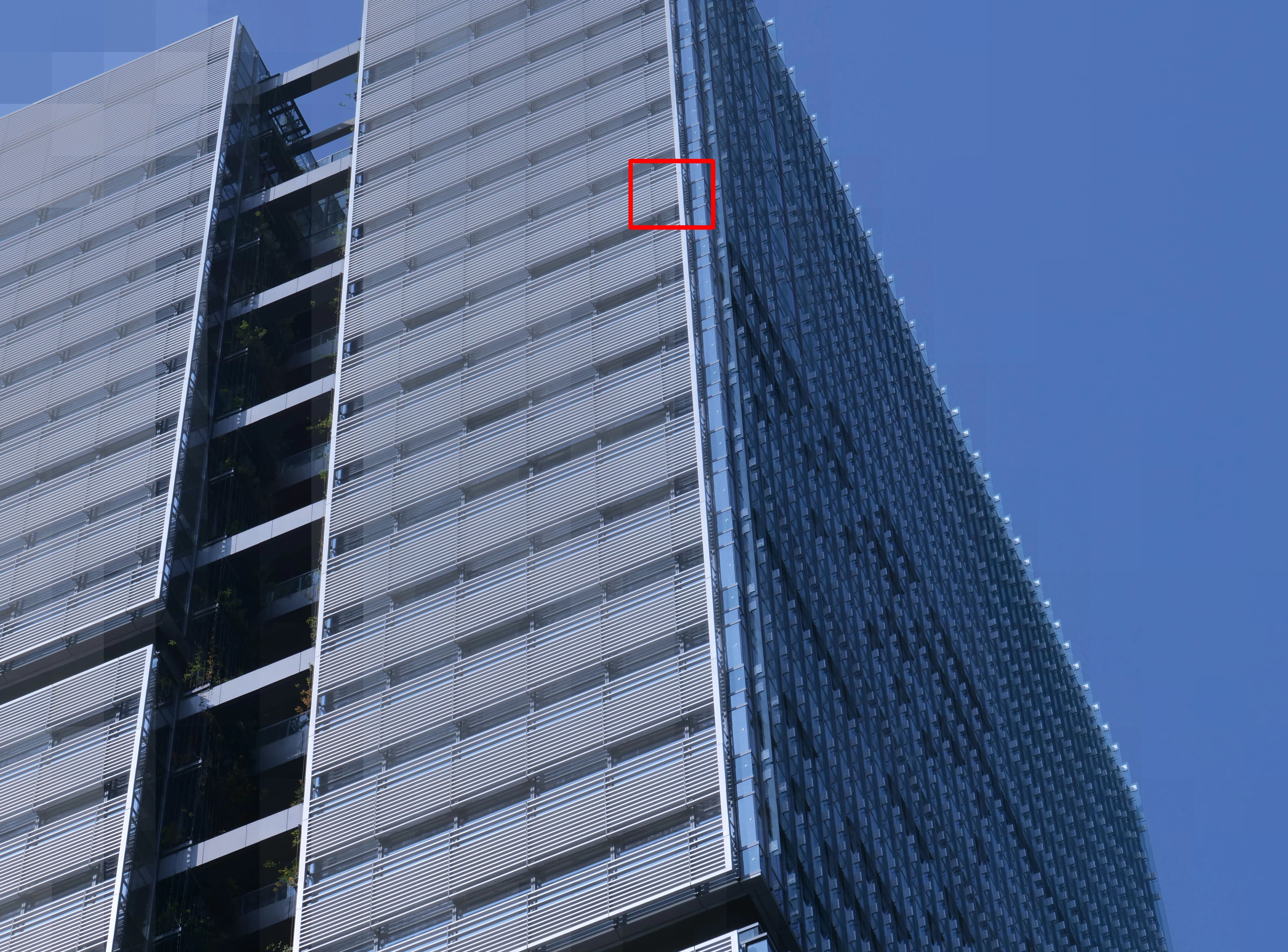}
						\\
						DRealSR ($4\times$): panasonic\_187
					\end{tabular}
				\end{adjustbox}
				\begin{adjustbox}{valign=t}
					\begin{tabular}{cccccc}
						\includegraphics[height=\h \textwidth, width=\w \textwidth]{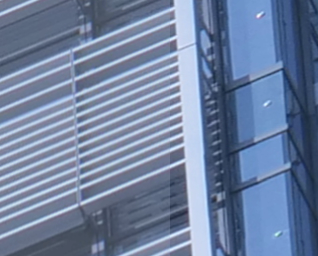} \hspace{\g} &
						\includegraphics[height=\h \textwidth, width=\w \textwidth]{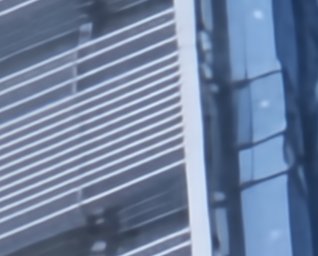} \hspace{\g} &
						\includegraphics[height=\h \textwidth, width=\w \textwidth]{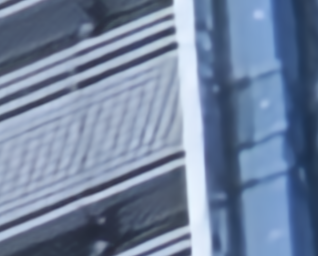} \hspace{\g} &
						\includegraphics[height=\h \textwidth, width=\w \textwidth]{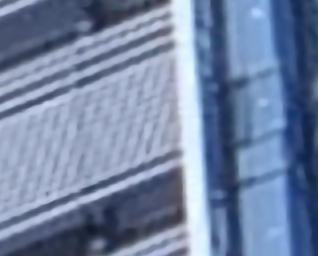} \hspace{\g} &
						\includegraphics[height=\h \textwidth, width=\w \textwidth]{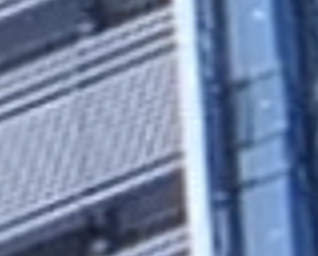} 
						\\
						HR \hspace{\g} &
						Target Only \hspace{\g} &
						Source Only \hspace{\g} &
						MZSR~\cite{MZSR} & 
                            ZSSR~\cite{ZSSR} \hspace{\g}

						\\
                            PSNR/SSIM \hspace{\g} &
                            28.43/0.8370 \hspace{\g} &
                            21.72/0.5894 \hspace{\g} &
                            20.35/0.4842 \hspace{\g} &
                            21.46/0.5311 \hspace{\g} 
                            \\
						\vspace{-2mm}
						\\

						\includegraphics[height=\h \textwidth, width=\w \textwidth]{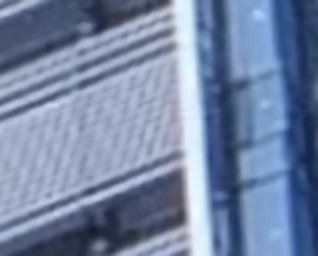} \hspace{\g} &
						\includegraphics[height=\h \textwidth, width=\w \textwidth]{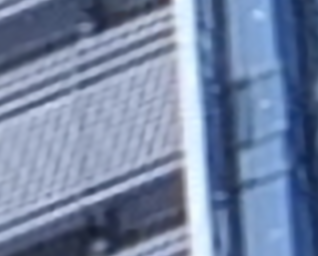} \hspace{\g} &
						\includegraphics[height=\h \textwidth, width=\w \textwidth]{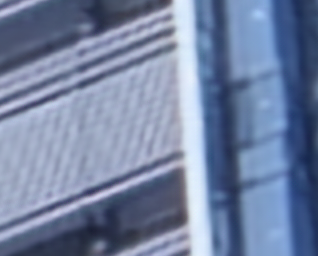}
						\hspace{\g} &		
						\includegraphics[height=\h \textwidth, width=\w \textwidth]{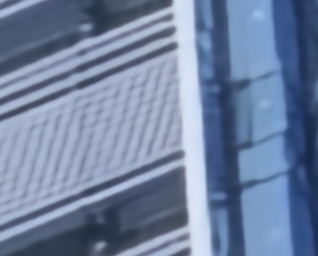} \hspace{\g} &
						\includegraphics[height=\h \textwidth, width=\w \textwidth]{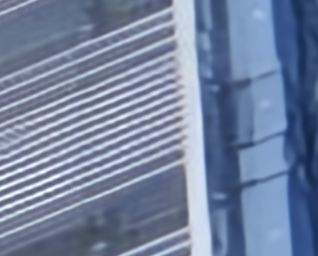} 
						\\ 
						
						CinCGAN~\cite{CinCGAN} \hspace{\g} &
						DASR~\cite{DASR}  \hspace{\g} &
						DRN-Adapt~\cite{DRN} \hspace{\g} &
						DADA~\cite{DADA}
						& \textbf{SODA-SR} (Ours)

						\\
                            21.40/0.5278 \hspace{\g} &
                            21.35/0.5295 \hspace{\g} &
                            21.81/0.5288 \hspace{\g} &
                            22.24/0.5670 \hspace{\g} &
                            25.58/0.7576 \\
					\end{tabular}
				\end{adjustbox}
			\end{tabular}
			
		\end{tabular}
	}
 
 \vspace{-2mm}
	\caption{Visual comparison for $\times$ 4 SR on DRealSR dataset (Sony$\rightarrow$Panasonic). Best viewed with zoom in.}
 \vspace{-0.3cm}
	\label{fig:visual_result}
\end{figure*}

Due to the absence of the source-free method for real-world image SR, we compare our SODA-SR with the state-of-the-art UDA methods and self-supervised methods for real-world image SR. The competing UDA methods include CinCGAN~\cite{CinCGAN}, DASR~\cite{DASR}, DRN-Adapt~\cite{DRN} and DADA~\cite{DADA} while self-supervised methods include ZSSR~\cite{ZSSR} and MZSR~\cite{MZSR}. Our experiments include real$\rightarrow$real adaptation and synthetic$\rightarrow$real adaptation. In the real$\rightarrow$real adaptation task, our source model is trained on the real-world image pairs from DRealSR. Alternatively, in the synthetic$\rightarrow$real adaptation task, our source model is trained on synthetic image pairs (\ie, LR images are bicubically downsampled from the real-world HR images). Our experiments are for scaling factor $\times 4$.

Table~\ref{tab:compare_uda} shows the quantitative results in six camera$\rightarrow$camera adaptation tasks. "Source only" represents the model trained on the source data without domain adaptation. "Target only" represents the model trained on the labeled target data. In six real$\rightarrow$real tasks, our method achieves the best performance on  PSNR and SSIM, and the second best performance on LPIPS. Although DASR achieves the best LPIPS, its PSNR and SSIM are inferior to ours, and it often produces SR images containing artifacts and noises. In six synthetic$\rightarrow$real tasks, our method also performs better than other methods.
Table~\ref{tab:compare_ssl} shows the quantitative results compared with self-supervised SR methods on three test sets. ZSSR and MZSR require only a single LR image to train a SR network specifically tailored to that LR image. Since our method preserves the domain-invariant knowledge in the pre-trained source model and reduces the cross-domain discrepancy by model adaptation, our method performs much better than the self-supervised SR methods.

Fig.~\ref{fig:visual_result} shows that our method can not only reason the correct structure of the buildings but also generate clear details, while other methods may generate deformed structure and blurry results. The results of our method are closest to those trained with target labels. To validate the effectiveness of the proposed method on other backbone network, we take experiments on using SwinIR~\cite{liang2021swinir} as the backbone. The results are reported in the Appendix for page limit.

\begin{table}[t]
\captionsetup{font=small}%
\scriptsize
\center
\begin{center}
\resizebox{\linewidth}{!}{
\begin{tabular}{c|c|cccc|c}
    \toprule
         Metrics & \makecell{Source\\ Only} & \makecell{w/o\\WAT} &\makecell{w/o\\EMA}&\makecell{w/o\\Reg. Loss}&\makecell{w/o\\UE}&Ours
         \\
         \midrule
         PSNR$\uparrow$ &30.49 & 30.64 & 31.14 & 31.31 &31.33 & \textbf{31.41}
         \\
         SSIM$\uparrow$ &0.820 & 0.822& 0.826 & 0.829 &0.830 & \textbf{0.832}
         \\LPIPS$\downarrow$ &0.363& 0.365 & 0.354 & 0.356 &0.346& \textbf{0.344}
         \\
         \bottomrule
    \end{tabular}}
\end{center}
    \vspace{-0.2cm}
    \caption{Ablation results on DRealSR (Olympus$\rightarrow$Panasonic). We evaluate the effectiveness of WAT, EMA strategy, regularization losses, and uncertainty estimation (UE).} 
    \vspace{-0.05cm}
    \label{tab:ablation}
    
\end{table}

\subsection{Ablation Study}
\begin{figure}
    \captionsetup{font=small}%
    \centering

    \includegraphics[width=0.47\textwidth]{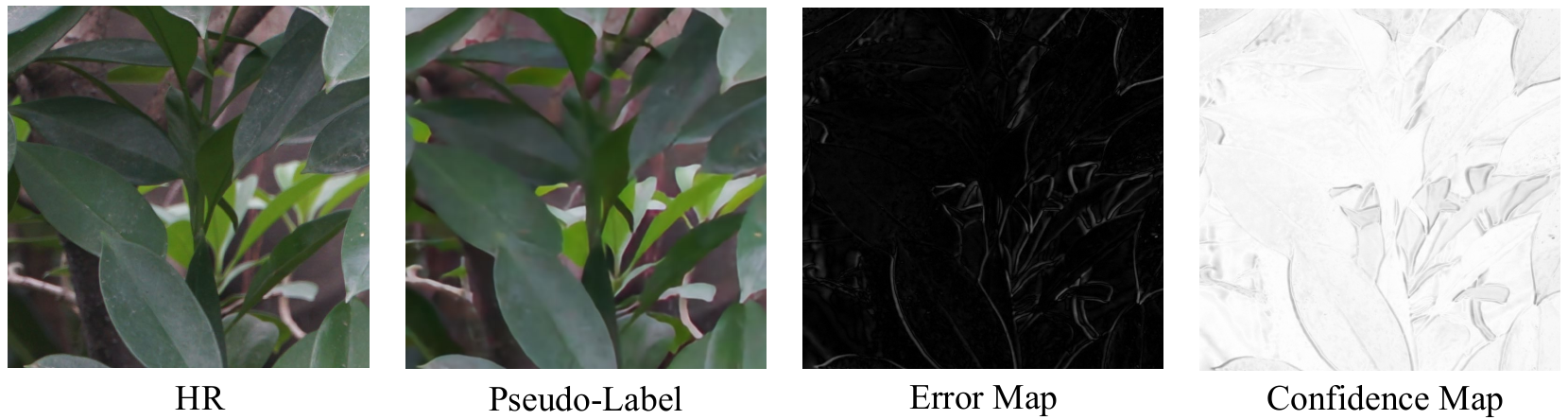}
    \vspace{-0.2cm}
    \caption{Visual illustration of the uncertainty estimation.}
    \vspace{-0.4cm}
    \label{fig:UE}
\end{figure}

\begin{figure}
   \captionsetup{font=small}%
    \centering
    \scriptsize
    \includegraphics[width=\linewidth]{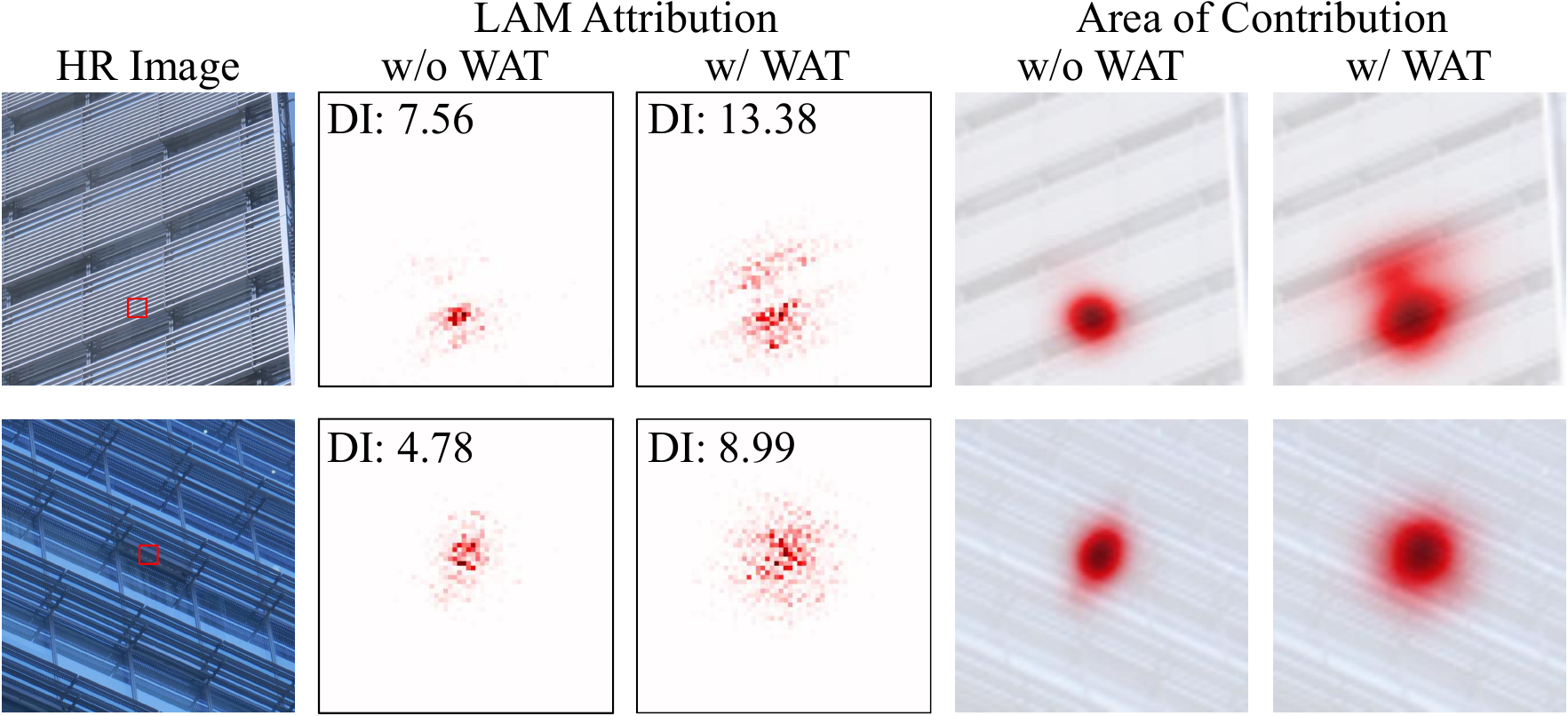}
    \vspace{-12pt}
    \caption{LAM~\cite{lam} illustration on challenging cases. LAM attribution shows how important each pixel in the LR image is for reconstructing the marked patch. Diffusion Index (DI) reflects how many pixels are used, with a higher DI meaning more pixels are involved.  The results indicate WAT enables the SR model to utilize a wider range of pixels for reconstruction.}
    \label{fig:ablation_WAT}
    \vspace{-0.6cm}
\end{figure}

We conduct an ablation study to evaluate the respective roles of each part in our method. As shown in Table~\ref{tab:ablation}, the performance drops a lot when separately removing WAT and EMA. This demonstrates the significant role of WAT and EMA in facilitating teacher-student learning. The results also demonstrate that the regularization loss and uncertainty estimation (UE) can improve the quality of SR images. As shown in Fig.~\ref{fig:UE}, pixels with higher error in the pseudo-label will be assigned lower confidence, which also proves the effectiveness of the proposed UE. Fig.~\ref{fig:ablation_WAT} also demonstrates that WAT can empower the model to utilize a broader range of pixels, leading to enhanced SR results. A detailed analysis of these components and the loss function in Eq.~(\ref{eq:loss_all}) are provided in the Appendix.

\section{Conclusion}
In this paper, we propose a novel source-free domain adaptation framework named SODA-SR, which attempts to adapt a source-trained model to the target domain without accessing source data. By using our proposed wavelet augmentation transformer (WAT), the student model is capable of learning low-frequency information of varying levels across diverse samples, which is aggregated efficiently via deformable attention. Besides, an uncertainty-aware self-training mechanism is proposed to facilitate knowledge transfer and rectify pseudo-labels. Several regularization losses are proposed to avoid overfitting pseudo-labels. Extensive experiments under real$\rightarrow$real and synthetic$\rightarrow$real adaptation settings on DRealSR demonstrate the effectiveness of our method. 

\vspace{1mm}

\noindent\textbf{Acknowledgements:} This research is partially funded by Youth Innovation Promotion Association CAS (Grant No. 2022132), Beijing Nova Program (20230484276), National Natural Science Foundation of China (Grant No. U21B2045, U20A20223) and OPPO Research fund.

{
    \small
    \bibliographystyle{ieeenat_fullname}
    \bibliography{main}
}


\end{document}